# Variance-Aware Adaptive Weighting for Diffusion Model Training

Nanlong Sun[1], Lei Shi[1],*

[1] Kennesaw State University, GA, USA

*Corresponding author: lshi@kennesaw.edu

**Abstract**

Diffusion models have recently achieved remarkable success in generative modeling, yet their training dynamics across different noise levels remain highly imbalanced, which can lead to inefficient optimization and unstable learning behavior. In this work, we investigate this imbalance from the perspective of loss variance across log-SNR levels and propose a variance-aware adaptive weighting strategy to address it. The proposed approach dynamically adjusts training weights based on the observed variance distribution, encouraging a more balanced optimization process across noise levels. Extensive experiments on CIFAR-10 and CIFAR-100 demonstrate that the proposed method consistently improves generative performance over standard training schemes, achieving lower Fréchet Inception Distance (FID) while also reducing performance variance across random seeds. Additional analysis, including loss-log-SNR visualization, variance heatmaps, and ablation studies, further reveal that the adaptive weighting effectively stabilizes training dynamics. These results highlight the potential of variance-aware training strategies for improving diffusion model optimization.

## 1. Introduction

Diffusion models have recently achieved remarkable success in image generation and related visual synthesis tasks [1-4,9-10,15-16,19]. Their training objective is typically formulated as a noise-conditioned denoising problem, where noise levels are sampled from a predefined distribution [1-2]. While architectural innovations and improved parameterizations have been extensively studied, the role of the noise-level sampling distribution in optimization dynamics remains less explored.

In common practice, diffusion models adopt heuristic noise sampling strategies such as log-uniform or log-normal distributions [3]. These fixed schedules implicitly determine how frequently different signal-to-noise ratio (SNR) regimes contribute to stochastic gradient updates. However, since training relies on Monte Carlo sampling over noise levels, the choice of sampling distribution directly influences the variance of the stochastic gradient estimator, which in turn affects convergence behavior and training stability [5,14].

In this work, we revisit diffusion training from the perspective of stochastic gradient variance [18]. By analyzing per-sample training loss statistics conditioned on log-SNR levels, we observe that the variance distribution across noise regimes is highly non-uniform. Particularly, certain intermediate SNR regimes contribute disproportionately to stochastic gradient variability. This observation suggests that commonly used fixed sampling strategies may lead to sub-optimal variance allocation during optimization.

To better understand this phenomenon, we derive a variance decomposition under log-SNR

sampling and connect diffusion training to classical variance-optimal importance sampling principles [6]. The analysis reveals that, in theory, the variance-minimizing sampling density should be proportional to the conditional standard deviation of gradients. Motivated by this insight, we propose a lightweight adaptive log-SNR reweighting strategy that approximates variance-aware importance sampling without modifying the underlying noise schedule [17].

The proposed method introduces a simple batch-level log-SNR weighting mechanism that flattens variance concentration across noise regimes. Importantly, it requires no architectural changes and incurs negligible computational overhead. Experiments on CIFAR-10 and CIFAR-100 demonstrate consistent improvements in FID over standard log-normal sampling, validating the practical benefits of variance-aware reweighting.

Our contributions can be summarized as follows:

- We empirically analyze gradient variance across log-SNR regimes in diffusion training.
- We establish a theoretical connection between log-SNR sampling and variance-optimal importance sampling.
- We propose a simple adaptive weighting strategy that improves generative performance without altering the noise schedule.

## 2. Related Work

### 2.1 Diffusion Models

Diffusion models have emerged as a powerful class of generative models for images synthesis [10-11]. The denoising diffusion probabilistic model (DDPM) introduced a discrete-time formulation that progressively adds and removes Gaussian noise during training and sampling [1]. Subsequent work connected diffusion modeling with score-based generative modeling under stochastic differential equations (SDEs), providing a continuous-time perspective and improved theoretical grounding [2].

Several improvements have been proposed to enhance training efficiency and sample equality [12]. For example, improved denoising diffusion probabilistic models introduced improved parameterizations and variance learning, while elucidating the design space of diffusion-based generative models (EDM) systematically analyzed design choices in diffusion training and proposed the EDM formulation [3-4]. These works primarily focus on architecture design, parameterization, and sampling acceleration.

In contrast, relatively less attention has been paid to understanding how the noise-level sampling distribution influences optimization dynamics during training.

### 2.2 Noise-Level Parameterization and Sampling Strategies

Noise-level parameterization plays a central role in diffusion training. Common practices adopt heuristic sampling distributions over noise scales, such as uniform, log-uniform, or log-normal sampling [3]. These strategies implicitly determine the frequency with which different signal-to-noise ratio (SNR) regimes contribute to stochastic gradient updates.

The log-SNR representation has been widely used to improve stability and interpretability in diffusion models [2]. However, existing works typically treat the sampling distribution as a fixed design choice rather than an optimization-sensitive component.

Our work revisits noise-level sampling from a gradient-variance perspective and provides empirical evidence that conditional training loss statistics exhibit strong heterogeneity across log-SNR levels.

### 2.3 Variance Reduction and Importance Sampling in Stochastic Optimization

Stochastic gradient descent (SGD) remains the backbone of diffusion training. The variance of stochastic gradient estimators plays a critical role in convergence behavior and training stability, as established in classical stochastic approximation theory [5]. Reducing gradient variance has long been recognized as an effective strategy for improving optimization efficiency.

Importance sampling provides a principled approach to variance reduction by allocating sampling probability proportional to the magnitude or variability of the integrand [6,13,20]. Variance-optimal sampling distributions are known to be proportional to the standard deviation of the target quality under mild conditions.

While importance sampling has been studied extensively in Monte Carlo estimation and stochastic optimization, its connection to log-SNR sampling in diffusion training has not been systematically explored. Our work establishes this connection and develops a practical reweighting strategy that approximates variance-aware importance sampling without modifying the underlying noise schedule.

### 3. Method

### 3.1 Problem Setup

Diffusion models are trained by minimizing a noise-conditioned denoising objective defined over a distribution of noise levels. Let $x \sim p_{data}(x)$ denote data samples and let $\sigma$ denote the noise scale. The general training objective can be written as

$$\mathcal{L}(\theta) = \mathbb{E}_{x,\sigma,\epsilon}[\ell(\theta; x, \sigma, \epsilon)] \qquad (1)$$

where $\sigma \sim p(\sigma)$ is the sampling distribution over noise levels and $\ell(\cdot)$ denotes the denoising loss under a specific parameterization (e.g., DDPM or EDM formulations).

During stochastic gradient descent, the expectation over $\sigma$ is approximated using Monte Carlo sampling. Consequently, the stochastic gradient estimator is

$$\nabla_\theta \mathcal{L}(\theta) = \mathbb{E}_{\sigma \sim p(\sigma)}[\nabla_\theta \ell(\theta; \sigma)] \qquad (2)$$

which shows that the noise sampling distribution $p(\sigma)$ directly affects the optimization dynamics.

To better expose the structure of noise-level sampling, we adopt the log-SNR parameterization commonly used in modern diffusion formulations [3]. Let

$$\lambda = logSNR(\sigma) \tag{3}$$

which induces a transformed sampling distribution $p(\lambda)$. Under this parameterization, the objective becomes

$$\mathcal{L}(\theta) = \mathbb{E}_{\lambda \sim p(\lambda)}[\ell(\theta; \lambda)] \tag{4}$$

From this perspective, diffusion training can be interpreted as minimizing an expectation over log-SNR levels. Importantly, the sampling density $p(\lambda)$ determines how frequently different noise regimes contribute to gradient updates.

Because stochastic optimization relies on a finite number of sampled noise levels, the variance of the gradient estimator depends on both the conditional trainng loss statistics and the sampling distribution. Therefore, the choice of $p(\lambda)$ is not merely a design preference but an optimization-critical factor that directly affects training stability and convergence.

As illustrated in Fig. 1, the adaptive weighting module is inserted after log-SNR sampling and directly modulates the training loss.

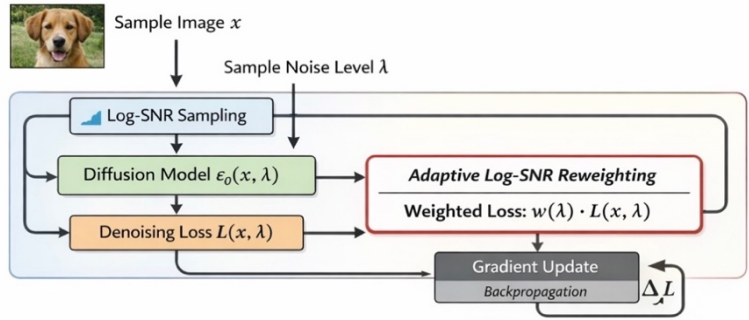

Fig. 1. Overview of the proposed adaptive log-SNR reweighting framework for diffusion training.

**3.2 Loss Variance Across Log-SNR Levels**

To understand how log-SNR affects optimization dynamics, we analyze training loss statistics conditioned on noise level. In practice, we measure the variance of per-sample training loss as a proxy for gradient variance.

Let $\ell(x, \lambda)$ denote the per-sample training loss under log-SNR level $\lambda$. We analyze the variance of per-sample losses conditioned on $\lambda$.

Using training samples collected during optimization, we estimate the conditional mean and conditional variance:

$$\mu(\lambda) = \mathbb{E}[G(x,\lambda)], \sigma^2(\lambda) = Var[G(x,\lambda)] \tag{5}$$

The statistics are computed by binning samples across the log-SNR domain at different training stages (10k, 30k, and 60k iterations).

**Mean Training Loss Magnitude.** We first analyze the conditional mean $\mu(\lambda)$. Empirically, the mean gradient magnitude remains relatively stable across the log-SNR domain and does not exhibit strong divergence at extreme noise levels.

**Conditional Loss Variance.** In contrast, the conditional variance $\sigma^2(\lambda)$ exhibits strong heterogeneity across log-SNR levels. Particularly, variance is concentrated in mid-to-high log-SNR regions and spans nearly two orders of magnitude in dynamics range. This pattern persists across training stages, indicating that it is intrinsic to the diffusion parameterization rather than a transient training artifact.

**Sampling Density Verification.** To ensure that the observed variance pattern is not caused by sampling imbalance, we also report the number of samples per log-SNR bin. Because the noise schedule is fixed during training, the sampling density remains stable across iterations. This confirms that the variance concentration arises from intrinsic training loss statistics rather than uneven sampling.

These observations suggest that certain log-SNR regions contribute disproportionately to stochastic gradient noise. Consequently, the interaction between the sampling distribution $p(\lambda)$ and conditional loss variance plays an important role in determining optimization efficiency.

**3.3 Variance-Optimal Importance Reweighting**

We now formalize the relationship between log-SNR sampling and stochastic gradient variance.

Let $g(x, \lambda) = \nabla_\theta \ell(x, \lambda)$ denote the per-sample gradient where $\lambda \sim p(\lambda)$. The variance of the stochastic gradient estimator can be decomposed using the law of total variance:

$$Var(g) = \mathbb{E}_\lambda[Var(g|\lambda)] + Var_\lambda(\mathbb{E}[g|\lambda]) \tag{6}$$

The first term corresponds to the within-log-SNR variance, while the second term captures variation of the conditional mean gradient across noise levels.

Let

$$\sigma^2(\lambda) = Var(g|\lambda) \tag{7}$$

The dominant contribution to stochastic noise in diffusion training is typically the first term:

$$\mathbb{E}_\lambda[\sigma^2(\lambda)] = \int p(\lambda)\sigma^2(\lambda)d\lambda \tag{8}$$

This expression reveals that the overall gradient variance is a weighted integral of conditional variance under the sampling distribution.

When $\sigma^2(\lambda)$ is highly non-uniform, fixed heuristic sampling strategies (e.g., log-normal sampling) may allocate samples inefficiently across noise levels.

**Variance-Optimal Sampling.** We consider the problem of minimizing gradient variance with respect to the sampling distribution.

Suppose the full gradient is estimated using Monte Carlo sampling:

$$\hat{g} = \frac{1}{N}\sum_{i=1}^{N} g(x_i, \lambda_i), \lambda_i \sim p(\lambda) \qquad (9)$$

The asymptotic variance of this estimator scales as

$$\int p(\lambda)\sigma^2(\lambda)d\lambda \qquad (10)$$

Minimizing this equality subject to $\int p(\lambda)d\lambda = 1$ yields the classical variance-optimal importance sampling distribution:

$$p^*(\lambda) \propto \sigma(\lambda) \qquad (11)$$

This result indicates that the optimal sampling probability should be proportional to the standard deviation of the integrand.

**Importance Reweighting Formulation.** Directly modifying the sampling distribution can be impractical in diffusion training, where the noise schedule is often coupled with model parameterization. Instead, importance sampling can be implemented equivalently through importance reweighting.

Given a base distribution $p(\lambda)$ and a target distribution $p^*(\lambda)$, an unbiased gradient estimator can be written as

$$\hat{g} = \frac{1}{N}\sum_{i=1}^{N} \frac{p^*(\lambda_i)}{p(\lambda_i)} g(x_i, \lambda_i), \lambda_i \sim p(\lambda) \qquad (12)$$

This formulation enables variance-aware optimization without modifying the underlying sampling process.

### 3.4 Adaptive Log-SNR Reweighting

Motivated by the variance heterogeneity observed in Section 3.2 and the variance-optimal sampling principle derived in Section 3.3, we propose a simple adaptive log-SNR reweighting mechanism.

Rather than altering the sampling distribution, we introduce a lightweight weighting function applied directly to the training loss.

Although the exact variance-optimal sampling distribution is proportional to the conditional gradient standard deviation, directly estimating this distribution during training is computational expensive.

Instead, we adopt a smooth adaptive weighting function centered around the batch-mean of log-

SNR values, which empirically approximates the variance-balancing behavior while maintaining computational simplicity.

For a mini-batch of samples with log-SNR values $\lambda$, we define the weight

$$w(\lambda) = exp(-\alpha(\lambda - \mu)^2) \tag{13}$$

where $\mu$ denotes the batch mean of log-SNR values and $\alpha$ controls the strength of reweighting.

This formulation attenuates the contribution of samples whose log-SNR deviates significantly from the batch center, effectively reducing the influence of regions associated with high conditional variance.

The resulting weighted loss becomes:

$$\ell_{weighted} = w(\lambda) \cdot \ell(\theta; x, \lambda) \tag{14}$$

This adaptive weighting strategy provides a practical approximation to variance-aware importance sampling while preserving the original noise schedule. As a result, the approach introduces negligible computational overhead and can be integrated into existing diffusion training pipelines without architecture modifications.

Empirical evaluations of the variance reduction behavior and its impact on generative performance are presented in Section 4.

## 4. Experiment

### 4.1 Experiment Setup

**Dataset.** Due to computational constraints, we focus on CIFAR-scale benchmarks. However, the proposed method is architecture-agnostic and can be directly extended to larger datasets. We conduct experiments on the CIFAR-10 and CIFAR-100 datasets, which contain 50,000 training images and 10,000 testing images at a resolution of $32 \times 32$ [7]. Datasets provide a diverse benchmark for evaluating generative models.

**Model and Training.** All models are trained within the EDM diffusion training framework using a standard U-Net architecture. Training is performed for 60k optimization steps using Adam optimizer with a learning rate of $2 \times 10^{-4}$ and batch size 128.

**Noise Sampling Baselines.** We compare out method with the commonly used noise sampling strategy: log-uniform sampling. Previous work shows that log-normal sampling consistently outperforms log-uniform strategies in EDM training [3]. Therefore, we adopt log-normal as the primary baseline.

Our approach introduces an adaptive variance-aware reweighting mechanism that adjusts the contribution of samples from different noise levels. Unless otherwise specified, the adaptive strength parameter is set to $\alpha = 0.05$.

**Evaluation Metric.** We evaluate generative quality using the Fréchet Inception Distance (FID),

computed from 50k generated samples [8].

**4.2 Main Results**

Table 1. FID Comparison Across Datasets.

| Method | FID | |
|---|---|---|
| | CIFAR-10 | CIFAR-100 |
| Log-normal | $14.21 \pm 0.31$ | $23.31 \pm 1.10$ |
| **Ours (adaptive)** | $\mathbf{13.58 \pm 0.55}$ | $\mathbf{20.89 \pm 0.74}$ |

Table 1 reports the generative performance of different sampling strategies on CIFAR-10 and CIFAR-100.

The log-uniform strategy leads to unstable optimization and produces significantly worse results. The proposed adaptive weighting approach achieves the best FID among all methods. Importantly, this improvement is obtained without modifying the model architecture or increasing computational cost. These results indicate that variance-aware training can effectively improve diffusion model optimization. In addition to improving the final FID score, the proposed method also reduces the variance across different training seeds, suggesting improved optimization stability (see Fig. 2).

**4.3 Loss Variance Analysis**

To better understand the effect of the proposed method, we analyze the distribution of per-sample training loss variance across different levels.

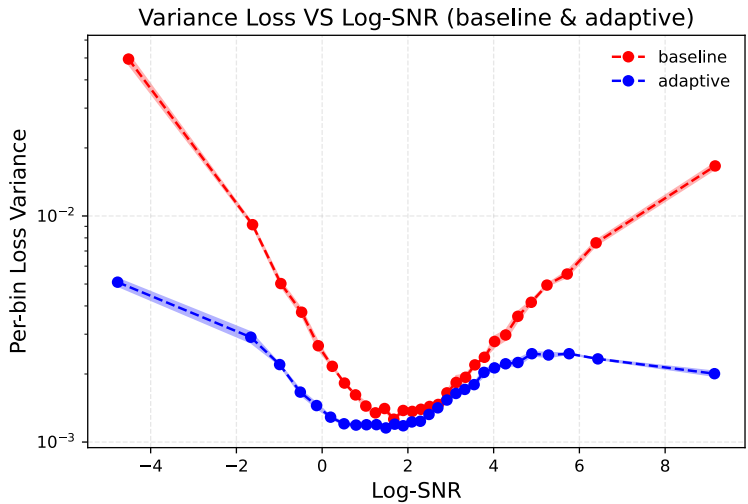

Fig. 2. Per-sample loss variance across log-SNR levels.

Specifically, we group training samples according to their log-SNR values and compute the variance of per-sample losses within each group. Fig. 2 illustrates the resulting variance distribution.

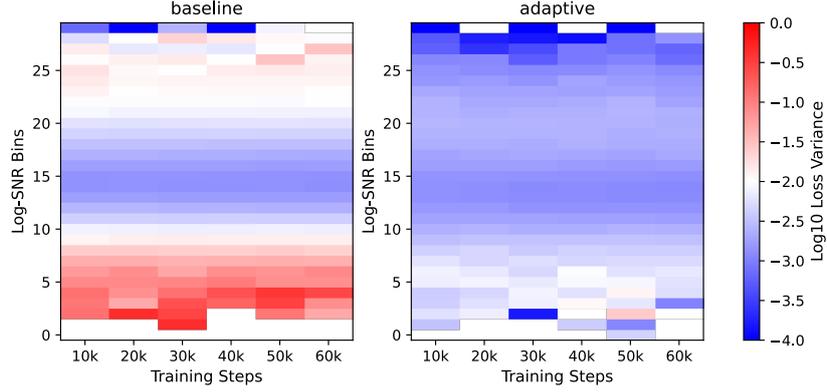

Fig. 3. Loss variance distribution across log-SNR bins during training.

The proposed adaptive weighting strategy results in a more balanced variance distribution compared to the baseline as shown in Fig. 3.

Under conventional sampling strategies, the loss variance distribution is highly uneven across noise levels. Certain log-SNR regions produce significantly larger loss variance, indicating that these regions dominate the training dynamics while others contribute less to optimization.

After applying the proposed adaptive weighting strategy, the loss variance becomes more balanced across log-SNR levels. This balanced distribution suggests that training signals are more evenly distributed, leading to improved optimization efficiency.

These observations indicate that reducing imbalance in per-sample loss variance can help stabilize diffusion model training.

**4.4 Visual Results**

To further evaluate the effectiveness of the proposed method, we present a qualitative comparison between the baseline diffusion model and our adaptive loss reweighting approach. The generated samples are shown in Fig. 4.

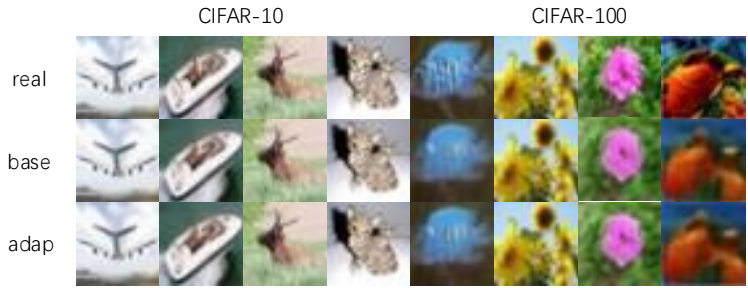

Fig. 4. Visual comparison on CIFAR-10 and CIFAR-100.

As illustrated in the Fig. 4, the samples produced by the baseline model occasionally exhibit artifacts and inconsistent structures. In contrast, the proposed method generates samples with improved visual coherence and more realistic details. This improvement can be attributed to the adaptive loss weighting mechanism, which balances the contribution of different noise levels during training and stabilizes the optimization process.

Overall, the visual results demonstrate that the proposed approach can produce higher-quality samples compared with the baseline diffusion objective.

### 4.5 Ablation Study

We further investigate the influence of the adaptive strength parameter $\alpha$. Three values are evaluated: $\alpha = 0.01$, $\alpha = 0.05$, and $\alpha = 0.1$.

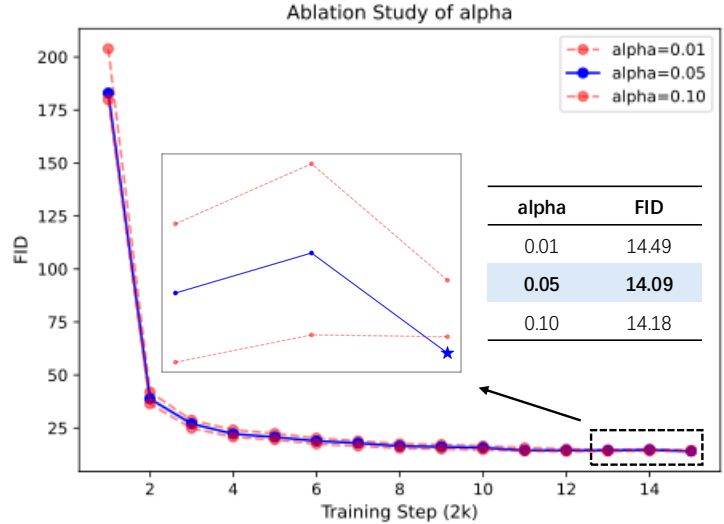

Fig. 5. Ablation study on the parameter alpha.

The Fig. 5 shows that small values introduce only mild reweighting effects, while overly large values may lead to excessive adjustment across noise levels. The best performance is achieved at $\alpha = 0.05$, which provides a good balance between stability and adaptability.

This ablation study confirms that moderate variance-aware reweighting is sufficient to produce consistent improvements.

### 4.6 Training Dynamics

Finally, we analyze the evolution of FID during training.

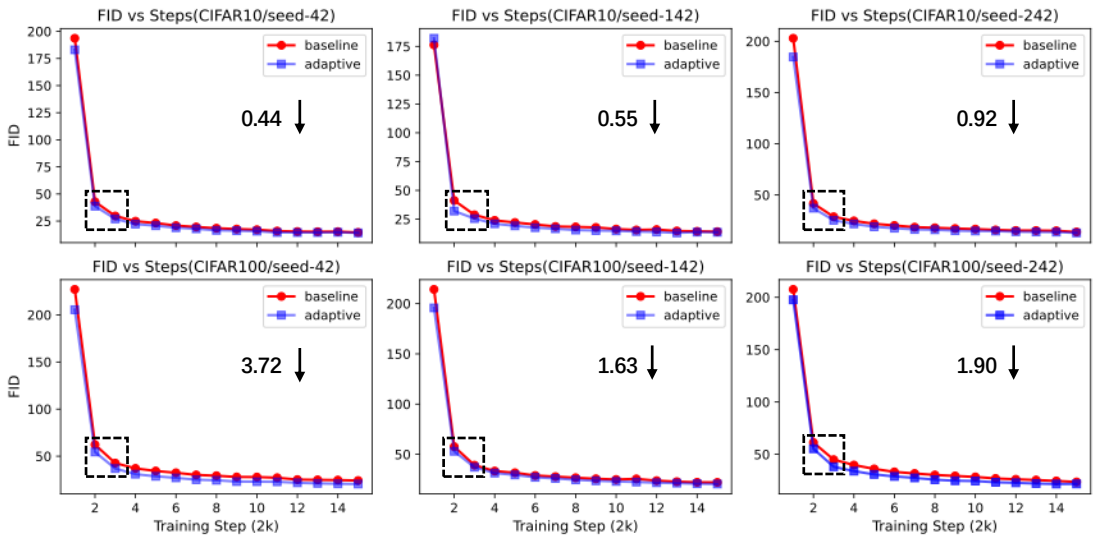

Fig. 6. FID versus training steps for baseline and the proposed adaptive method under three different random seeds on CIFAR-10 and CIFAR-100.

The Fig. 6 shows the FID curves across training iterations. Compared with baseline sampling strategies, the proposed method achieves faster convergence and maintains a lower FID throughout training.

These results demonstrate that adaptive variance-aware reweighting not only improves the final model quality but also accelerates training convergence.

## 5. Conclusion

In this work, we investigated the imbalance of training dynamics across different noise levels in diffusion models and proposed an adaptive weighting strategy to alleviate this issue. By adjusting the training weights according to the variance of the loss across log-SNR levels, the proposed method encourages a more balanced optimization process throughout training.

Extensive experiments on CIFAR-10 and CIFAR-100 demonstrate that the proposed adaptive strategy consistently improves generation quality compared with standard training schemes. Particularly, the method achieved lower FID scores while also reducing variance across different random seeds, indicating improved training stability. Additional analysis, including loss-log-SNR visualization, variance heatmaps, and ablation studies, further confirm that the adaptive weighting effectively balances the training dynamics across noise levels.

Overall, the results suggest that variance-aware adaptive weighting provides a simple yet effective mechanism for improving diffusion model training. The proposed framework is lightweight and can be readily integrated into existing diffusion pipelines. Future work may explore extending this strategy to larger datasets and more complex diffusion architectures, as well as investigating other adaptive criteria for balancing training dynamics.

**Acknowledgements**

The authors thank colleagues and reviewers for helpful discussions and suggestion.

**Reference**

[1] Ho J, Jain A, Abbeel P. Denoising diffusion probabilistic models[J]. Advances in neural information processing systems, 2020, 33: 6840-6851.

[2] Song Y, Sohl-Dickstein J, Kingma D P, et al. Score-based generative modeling through stochastic differential equations[J]. arXiv preprint arXiv:2011.13456, 2020.

[3] Karras T, Aittala M, Aila T, et al. Elucidating the design space of diffusion-based generative models[J]. Advances in neural information processing systems, 2022, 35: 26565-26577.

[4] Nichol A Q, Dhariwal P. Improved denoising diffusion probabilistic models[C]//International conference on machine learning. PMLR, 2021: 8162-8171.

[5] Needell D, Srebro N, Ward R. Stochastic gradient descent, weighted sampling, and the randomized Kaczmarz algorithm[J]. Advances in neural information processing systems, 2014, 27.

[6] Owen A. Lattice sampling revisited: Monte Carlo variance of means over randomized orthogonal arrays[J]. The Annals of Statistics, 1994: 930-945.

[7] Krizhevsky A, Hinton G. Learning multiple layers of features from tiny images[J]. 2009.

[8] Heusel M, Ramsauer H, Unterthiner T, et al. Gans trained by a two time-scale update rule converge to a local nash equilibrium[J]. Advances in neural information processing systems, 2017, 30.

[9] Dhariwal P, Nichol A. Diffusion models beat gans on image synthesis[J]. Advances in neural information processing systems, 2021, 34: 8780-8794.

[10] Li J, Shen T, Gu Z, et al. Adm: Accelerated diffusion model via estimated priors for robust motion prediction under uncertainties[C]//2024 IEEE 27th International Conference on Intelligent Transportation Systems (ITSC). IEEE, 2024: 2221-2227.

[11] Rombach R, Blattmann A, Lorenz D, et al. High-resolution image synthesis with latent diffusion models[C]//Proceedings of the IEEE/CVF conference on computer vision and pattern recognition. 2022: 10684-10695.

[12] Livni R, Shalev-Shwartz S, Shamir O. On the computational efficiency of training neural networks[J]. Advances in neural information processing systems, 2014, 27.

[13] Gower R M, Schmidt M, Bach F, et al. Variance-reduced methods for machine learning[J]. Proceedings of the IEEE, 2020, 108(11): 1968-1983.

[14] Song Y, Dhariwal P, Chen M, et al. Consistency models[J]. 2023.

[15] Nichol A, Dhariwal P, Ramesh A, et al. Glide: Towards photorealistic image generation and editing with text-guided diffusion models[J]. arXiv preprint arXiv:2112.10741, 2021.

[16] Saharia C, Chan W, Saxena S, et al. Photorealistic text-to-image diffusion models with deep language understanding[J]. Advances in neural information processing systems, 2022, 35: 36479-36494.

[17] Ho J, Salimans T. Classifier-free diffusion guidance[J]. arXiv preprint arXiv:2207.12598, 2022.

[18] Bottou L. Stochastic gradient learning in neural networks[J]. Proceedings of Neuro-Nımes, 1991, 91(8): 12.

[19] Lin S, Liu B, Li J, et al. Common diffusion noise schedules and sample steps are flawed[C]//Proceedings of the IEEE/CVF winter conference on applications of computer vision. 2024: 5404-5411.

[20] Bottou L, Curtis F E, Nocedal J. Optimization methods for large-scale machine learning[J]. SIAM review, 2018, 60(2): 223-311.

## Appendix A. Implementation Details

All models are trained by using Adam optimizer with a learning rate of $2 \times 10^{-4}$ and a batch size of 128. Training is performed for 60k iterations. We adopt the standard U-Net architecture (see Appendix B) used in diffusion models, following the configuration of Karras et al [3].

Noise levels are parameterization using log-SNR. Baseline sampling strategies is log-normal distribution over log-SNR, while our adaptive method dynamically adjusts sampling probabilities based on the estimated loss variance.

Model performance is evaluated using Fréchet Inception Distance (FID), computed with 50k generated samples. All experiments are conducted on a single NVIDIA RTX 4090 GPU and repeated with three random seeds.

## Appendix B. Network Architecture

The denoising network used in our experiments is based on a U-Net backbone. The model adopts a symmetric encoder-decoder structure with skip connections to facilitate multi-scale feature propagation. In the encoder, the spatial resolution is progressively reduced through down-sampling layers, while the number of feature channels increases. Residual blocks are employed at each stage to improve optimization stability. Attention modules are introduced at deeper layers to capture long-range dependencies. In the decoder, feature maps are gradually up-sampled and fused with the corresponding encoder features through skip connections. Additionally, time embeddings derived from the noise level are injected into each residual block to condition the network during diffusion training. Additional architectural details are illustrated in Fig. A1.

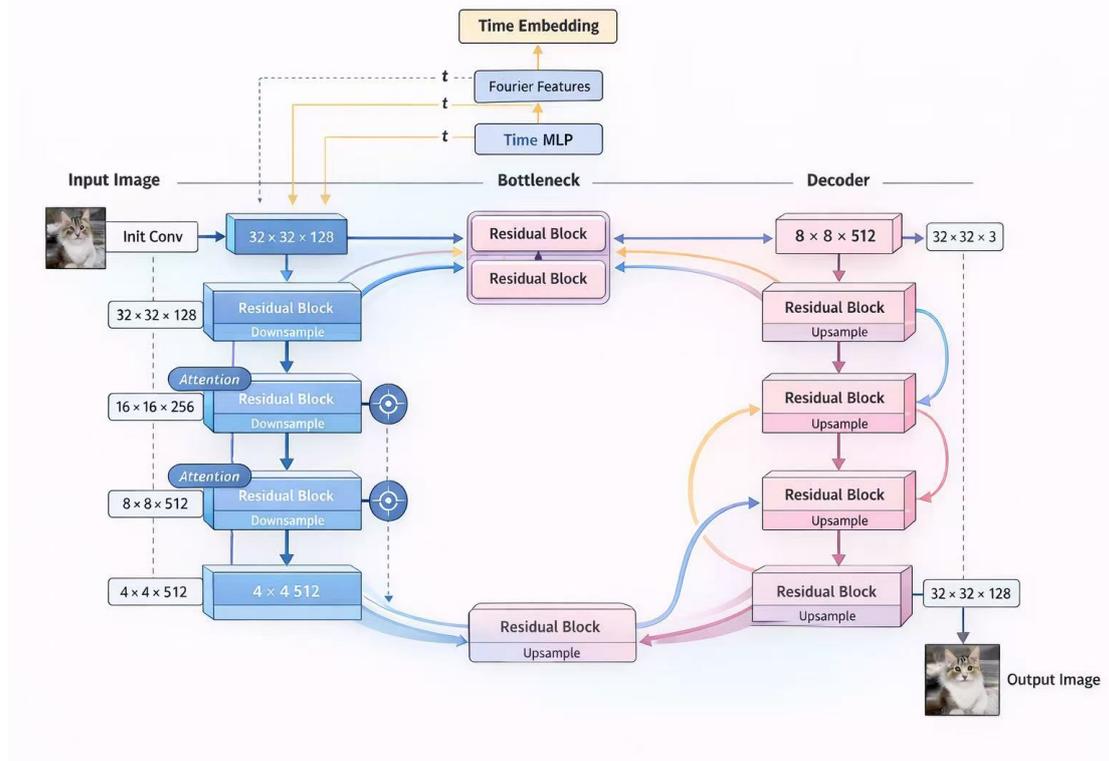

Fig. A1. Architecture of the U-Net backbone used in our diffusion model.

**Appendix C. Training Stability**

To verify that the proposed adaptive log-SNR weighting strategy does not negatively affect the optimization process, we compare the training loss curves of the baseline EDM objective and our weighted object during training on CIFAR-10 and CIFAR-100.

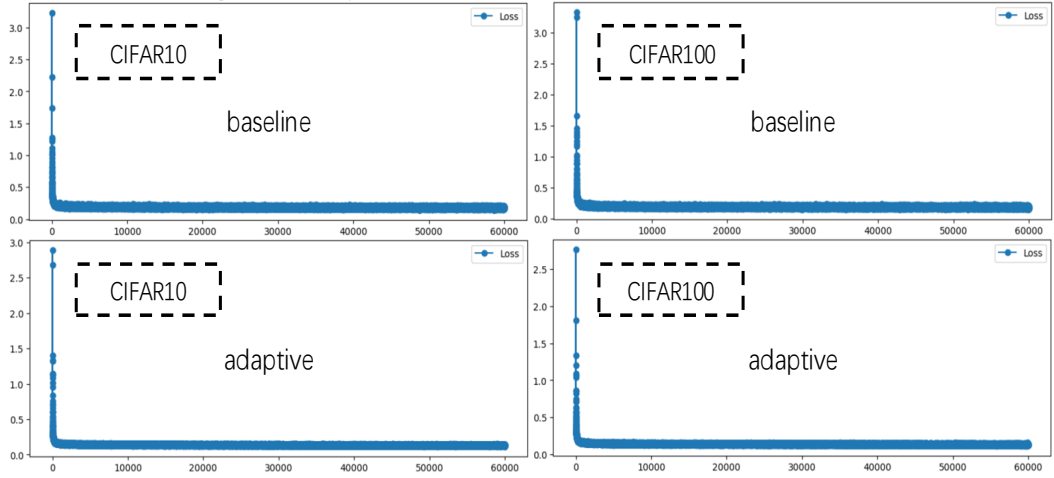

Fig. A2. Training loss curves of baseline EDM objective and the proposed adaptive log-SNR weighting objective on CIFAR-10 and CIFAR-100.

Fig. A2 illustrates the evolution of the training loss as a function of the training steps. As shown in the figure, both methods exhibit stable optimization behavior and converge smoothly throughout the training process. Importantly, the introduction of the adaptive log-SNR weighting does not introduce noticeable instability, oscillation, or divergence during training.

**Appendix D. Adaptive Log-SNR Reweighted EDM Loss (ALSR)**

In this section, we provide the detailed formulation of the training objective used in our method.

Our approach builds upon the EDM training framework, where the model learns to denoise a noisy image generated by adding Gaussian noise with a sampled noise level $\sigma$. To improve the training stability across different noise regimes, we introduce an adaptive reweighting mechanism based on the log-SNR.

Specifically, we compute the log-SNR corresponding to each sampled noise level and apply a variance-aware weighting term that down-weights extreme noise regimes. The overall training objective combines the standard EDM weighting with the proposed log-SNR adaptive factor.

The complete training procedure is summarized in Algorithm A1.

**Algorithm 1. Adaptive log-SNR Reweighted EDM (ALSR)**

| | |
|---|---|
| 1 | Sample noise level $\sigma \sim p(\sigma)$ |
| 2 | Sample Gaussian noise $\varepsilon \sim N(0, I)$ |
| 3 | $\tilde{x} = x + \sigma\varepsilon$ |
| 4 | Compute EDM coefficients: |
| 5 | $$c_{skip} = \sigma\_data^2/(\sigma^2 + \sigma\_data^2)$$ |
| 6 | $$c_{out} = \sigma \cdot \sigma_{data}/\sqrt{\sigma^2 + \sigma\_data^2}$$ |
| 7 | $$c_{in} = 1/\sqrt{\sigma^2 + \sigma\_data^2}$$ |
| 8 | $r = f_\theta(c_{in} \cdot \tilde{x}, \sigma)$ |
| 9 | $D(\tilde{x}, \sigma) = c_{skip} \cdot \tilde{x} + c_{out} \cdot r$ |
| 10 | $w_{EDM} = (\sigma^2 + \sigma\_data^2)/(\sigma \cdot \sigma\_data)^2$ |
| 11 | $s = log(\sigma\_data^2/\sigma^2)$ |
| 12 | $w_{SNR} = 1/(1 + \alpha(s - \mu)^2)$ |
| 13 | $L = w_{EDM} \cdot w_{SNR} \cdot \|D(\tilde{x}, \sigma) - x\|^2$ |

## Appendix E. Code Availability

To facilitate reproducibility, the implementation of the proposed method will be publicly released on GitHub upon acceptance of the paper.